\documentclass[10pt, journal]{IEEEtran}
\usepackage{amsmath,amssymb,amsfonts}
\usepackage{graphicx}
\usepackage{pifont}

\RequirePackage{silence}
\usepackage{amsthm}
\usepackage{xcolor}
\usepackage[figuresright]{rotating}

\usepackage{graphicx}
\graphicspath{{/}{fig/}}

\usepackage{array}
\usepackage{textcomp}
\usepackage{xcolor}
\usepackage{multirow}
\usepackage{booktabs}
\usepackage{enumerate}

\usepackage{mathtools}
\usepackage{breqn}
\usepackage{float}

\usepackage{pgfplots}
\pgfplotsset{compat=1.7}
\usepgfplotslibrary{groupplots}

\usepackage[font=small]{caption}
\usepackage{subcaption}

\usepackage{multirow,tabularx}
\usepackage{hyperref}
\usepackage{flushend}
\usepackage{algorithmic}

\usepackage[vlined, ruled, shortend]{algorithm2e}

\newlength\figureheight
\newlength\figurewidth
\SetAlCapNameFnt{\footnotesize}
\SetAlCapFnt{\footnotesize}

\title{

    Exploiting Redundancy for UWB Anomaly Detection in Infrastructure-Free Multi-Robot Relative Localization

}

\author{
    \IEEEauthorblockN{
        \vspace{1em}
        Sahar Salimpour\IEEEauthorrefmark{1},
        Paola Torrico Mor\'on\IEEEauthorrefmark{1},
        Xianjia Yu\IEEEauthorrefmark{1},
        Tomi Westerlund\IEEEauthorrefmark{1},
        Jorge Pe\~na Queralta\IEEEauthorrefmark{1} \\[+0.85em]
    }
    \IEEEauthorblockA{
        \normalsize
        \IEEEauthorrefmark{1}\href{https://tiers.utu.fi}{Turku Intelligent Embedded and Robotic Systems (TIERS) Lab, University of Turku, Finland}.\\
        Emails: \textsuperscript{1}\{sahars, pctomo, xianjia.yu, jopequ, tovewe\}@utu.fi
    }
}

\begin{document}

\maketitle
\thispagestyle{empty}
\pagestyle{empty}

%%%%%%%%%%%%%%%%%%%%%%%%%%%%%%%%%%%%%%%%%%%%%%
%%                                          %%
%%           ABSTRACT AND TITLE             %%
%%                                          %%
%%%%%%%%%%%%%%%%%%%%%%%%%%%%%%%%%%%%%%%%%%%%%%

\begin{abstract}\label{sec:abstract}%
Ultra-wideband (UWB) localization methods have emerged as a cost-effective and accurate solution for GNSS-denied environments. There is a significant amount of previous research in terms of resilience of UWB ranging, with non-line-of-sight and multipath detection methods. However, little attention has been paid to resilience against disturbances in relative localization systems involving multiple nodes. This paper presents an approach to detecting range anomalies in UWB ranging measurements from the perspective of multi-robot cooperative localization. We introduce an approach to exploiting redundancy for relative localization in multi-robot systems, where the position of each node is calculated using different subsets of available data. This enables us to effectively identify nodes that present ranging anomalies and eliminate their effect within the cooperative localization scheme. We analyze anomalies created by timing errors in the ranging process, e.g., owing to malfunctioning hardware. However, our method is generic and can be extended to other types of ranging anomalies. Our approach results in a more resilient cooperative localization framework with a negligible impact in terms of the computational workload.
\end{abstract}

\begin{IEEEkeywords}
% Keywords
multi-robot systems; ultra-wideband (UWB) localization; anomaly detection; positioning; relative localization
\end{IEEEkeywords}
\IEEEpeerreviewmaketitle

%%%%%%%%%%%%%%%%%%%%%%%%%%%%%%%%%%%%%%%%%%%%%%
%%                                          %%
%%                secS                      %%
%%                                          %%
%%%%%%%%%%%%%%%%%%%%%%%%%%%%%%%%%%%%%%%%%%%%%%

\section{Introduction}\label{sec:introduction}

In recent years, there has been a growing interest in the development of anomaly detection in various applications, including fraud and fault detection in safety systems~\cite{yang2017anomaly}, healthcare~\cite{fernando2021deep}, and autonomous robots~\cite{salimpour2022self}. Anomaly detection, also known as foreign detection or outlier detection, is the process of finding anomalous patterns in a given dataset. It can be applied to different types of data, including numerical data, images, videos, audio, text, or time series data~\cite{bogdoll2022anomaly, chatterjee2022iot}. Many multi-robot applications are also at risk of being affected by anomalous data or byzantine agents which can disrupt the entire operation~\cite{ferrer2021following, deng2021investigation}. The attack can target visual sensors, communication sensors, and positioning and localization sensors such as lidars, GNSS, or ultra-wideband (UWB). In our recent work, we presented a decentralized method for anomalies and byzantine agent detection in multi-robot systems using an external motion capture (MOCAP) system to determine the robot's location~\cite{salimpour2022decentralized}. In this paper, our objective is to use UWB technology for multi-robot relative localization while being able to detect byzantine robots that generate false or anomalous ranging information. 

\begin{figure}[t]
    \centering
    \includegraphics[width=.49\textwidth]{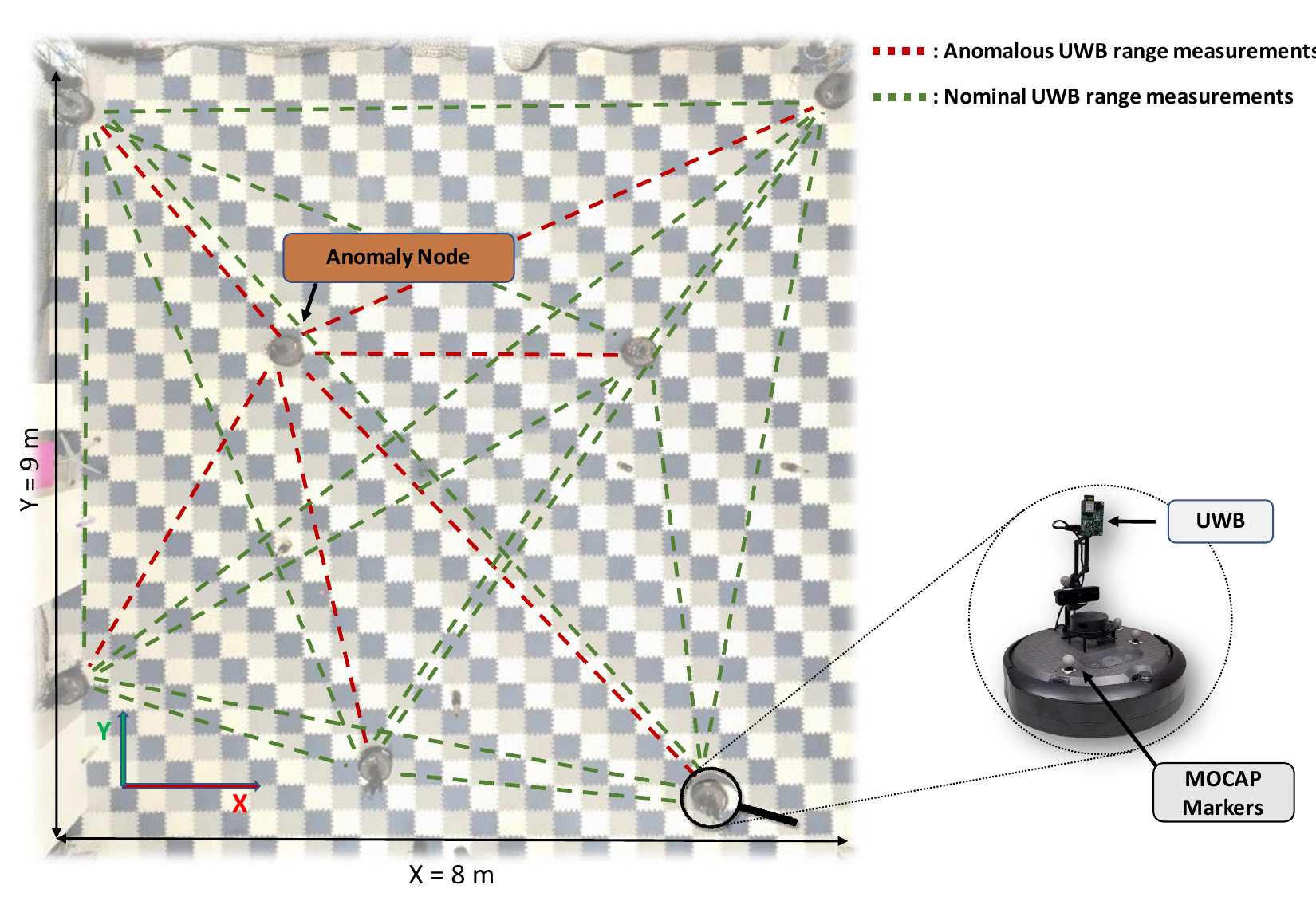}
    \caption{Illustration of the experimental arena and customized Turtlebot\,4 robots used in the experiments. Here we show in red the range measurements that are inaccurate, with nominal ranges in green.}
    \label{fig:experimental_setup}
\end{figure}

Global Navigation Satellite Systems (GNSS), such as the Global Positioning System (GPS), are widely employed for accurate global positioning in outdoor environments. Nonetheless, GNSS cannot reliably perform indoors and may be disrupted by attacks, or by jamming signals at the receiver. Among the localization approaches in GNSS-denied and indoor environments, UWB technology has emerged as a robust solution for relative localization and state estimation in multi-robot systems~\cite{queralta2020uwb}. A UWB-based system could replace expensive and complicated MOCAP systems and achieve centimeter-level localization accuracy for mobile robots. Fixed anchors and mobile tags are the two types of nodes used in UWB localization, while time of flight (ToF) and time difference of arrival (TDoA) are two popular methods for ranging measurements~\cite{moron2022towards}.

 % Note: Talk more about postioning, generally explain positioing algorithms and what are differences between them 
In this paper, a multi-robot cooperative positioning system employs UWB nodes in each robot to achieve robust relative localization in GNSS-denied environments through information sharing. For UWB ranging, we rely on time-of-flight (ToF) measurements between all pairs of nodes. Trilateration or multilateration algorithms can be used to determine the position of the tags~\cite{wang2021denoising}. In 2D localization, which relies on at least two nodes to determine the tag's position, the estimated positions might slightly vary depending on the reference nodes used. Using all available nodes, we propose a multiconfigurational localization method to determine optimal positions and make them resilient to fixed anchors. The core idea is to exploit redundancy in terms of the ways in which the relative positions can be calculated. This is possible because a higher number of ranges are measured than the minimum needed to ensure a unique solution in the relative localization problem (except for transaltions, rotations and mirror solutions).

Numerous research has been conducted in the UWB relative localization systems in terms of obtaining robust and reliable localization performance in swarms of robots. However, wireless networking makes multi-robot systems vulnerable to various types of attacks, so the UWB ranging mechanism is not immune to ranging attacks, such as malicious interference, sensor failure, timing error, or jamming, among others. In multi-robot cooperative localization, it is crucial not only to detect the anomalous pattern but also to identify and eliminate byzantine robot that can tamper with their ToF measurement, manipulating the distance measurement and creating false ranging information for other nodes~\cite{karapistoli2014adlu}.

Statistical-based approaches in related works, such as Kalman filters~\cite{ma2019robust}, detect outliers and estimate normal values in UWB ranges. However, in this study, the anomaly is a UWB transceiver (robot) that may generate a normal but altered data pattern. We propose a robust anomaly detection framework based on the fusion of all possible position configurations using different reference nodes. We developed our code based on ROS\,2 and verified the proposed approach in real-world robot navigation. Our proposed method identifies unreliable UWB node at each time stamp and eliminates them from the cooperative localization pipaline within the multi-robot system. In this contribution, we calculate the least square error of the reference node's corresponding configurations and the dispersion of the remaining configurations after applying fixed-UWB range matching. 

The remainder of this document is organized as follows. In Section~II, we review related works for UWB relative localization and anomaly detection in robotics. Section~III describes our methodology for cooperative localization and the detection of abnormal UWB data and robots. Section~IV describes the experimental results. Finally, Section~V summarizes the work and outlines future directions.

\section{Related Work}\label{sec:related_work}

\subsection{Anomaly Detection in Robotics}

Detection of anomalies in robotics, such as unusual data patterns~\cite{he2018admost}, abnormal behaviors, or abnormal sensors, can broadly be divided into two areas: self-monitoring and group monitoring. In self-monitoring approaches, each robot detects anomalies independently. Meanwhile in group-detection approach, multiple robots are involved in detecting malicious data or byzantine agents~\cite{salimpour2022decentralized}. In both scenarios, knowledge-based approaches typically learn the distribution of normal data in order to make the model robust to abnormal data. In~\cite{olive2020detection}, a simple autoencoder framework based on reconstruction error was presented to detect anomalies in trajectory data. Such approaches require sufficient samples for the training data.

\subsection{UWB-based Localization in Mobile Robots}

UWB has been widely adopted in robotics and autonomous field and its applications include global or relative localization and wireless mesh sensor networks for situated communication~\cite{xianjia2021applications}. UWB can significantly benefit  localization in the robotics field, particularly in GNSS-denied environments~\cite{alarifi2016ultra}, seamlessly transition between indoor and outdoor environments while keeping centimeter-level global or relative position accuracy~\cite{almansa2020autocalibration}. Most of the UWB positioning solutions, both commercial and in academia, are based on fixed anchors in known locations. Some studies are focusing on utilizing various approaches to improve localization accuracy. In~\cite{poulose2020uwb}, the authors applied Long Short-term Memory (LSTM) to estimate the user position based on anchors and achieved comparable accuracy to traditional approaches including triangulation. More recently, relative localization has gained traction within multi-robot collaborative localization problems as it enables higher degrees of flexibility and infrastructure-free deployments~\cite{xianjia2023loosely}. With multiple UWB transceivers mounted in each robot, relative positions among robots can be estimated ~\cite{nguyen2018robust} for various applications including autonomous docking~\cite{nguyen2019integrated}, collaborative scene reconstruction~\cite{queralta2022vio}, and others. By integrating UWB with other sensors including visual odometry, and GNSS, the robots can obtain more accurate and robust relative positions~\cite{xu2020decentralized, qi2020cooperative, xianjia2021cooperative}.

\subsection{UWB Anomaly Detection}

Outliers and malicious data are crucial factors that can significantly affect the accuracy of UWB-based positioning algorithms. To boost the accuracy and reliability of localization, many researchers have proposed various solutions to detect and address outlier data in both line-of-sight (LOS) and non-line-of-sight (NLOS) conditions. For example, the authors in~\cite{dwek2019improving} used an extended Kalman filter (EKF) for sensor fusion, and outlier detection based on residual error on UWB ranges at each time step. In~\cite{che2021anomaly}, an anomaly detection approach is presented for UWB indoor positioning based on Gaussian Distribution (GD) and Generalized Gaussian Distribution (GGD) algorithms. This paper has used the variance of the estimated distance and the power of the first path to classify NLoS environment. In another study~\cite{wang2021denoising}, authors proposed an anomaly detection treatment using the derivatives of the mean function of a UWB range time series to denoise and correct dropouts and outliers.

In summary, most existing approaches look for out-of-distribution data. Such statistical outlier detection methods are not ideal for anomalous agent detection problems, as large numbers of data need to be analyzed with a high computational load and may not detect faults immediately. For example, individual range distributions might be nominal to a LOS measurement but shifted or biased. To approach this issue, we present instead an approach where we exploit redundancy in ranging measurements and perform a statistical outlier detection at the multi-robot system level rather than invidudal measurement level.
\section{Methodology}\label{sec:methodology}

This section presents details of the multi-robot cooperative localization approach we present for an infrastructure-free (or anchor-free) anomaly-resilient relative localization. We also compare the anomaly detection process with multilateration to the use of a gradient descent algorithm.

\subsection{Hardware}

In the experiments, we used eight TurtleBot4 Lite mobile robot platforms built on top of the iRobot Create 3 base, denoted as $\mathcal{R}_{i,\:i\in\{1,2,3,4,5,6,7,8\}}$. Each of these robots was outfitted with a Qorvo DWM1001 UWB transceiver that had custom firmware installed. We developed and deployed custom firmware for the DWM1001 modules, enabling them to function as active nodes. To gather ranging measurements, we programmed the UWB transceivers to repeatedly measure the ToF between each pair of nodes. Within the experimental arena, the robots move around while an OptiTrack motion capture (MOCAP) system provides accurate ground truth information to verify the estimated relative states of each robot. The operating area is approximately 8 meters wide, 9 meters long, and 5 meters high (see Fig.~\ref{fig:experimental_setup}). Within this area, we positioned two static nodes, $\mathcal{R}_{1}$ and $\mathcal{R}_{2}$, and six mobile robots, denoted as $\mathcal{R}_{i,\:i\in\{3,4,5,6,7,8\}}$. The purpose of the static nodes is to provide a common orientation reference for comparing different relative localization approaches.

\begin{algorithm}[t]

    \small
	\caption{cooperative localization}
	\label{alg:Collabrative localization}
	\KwIn{\\
            \hspace{1em}Number of UWB nodes: \textit{{N}};\\
	    \hspace{1em}UWB ToF Ranges: ${d}_{(i,j)}$;
	}
    \KwOut{\\
      \hspace{1em}Robot poses
    }
    \SetKwFunction{Flse}{minimum\_shift\_lse}
    \SetKwFunction{Fmatch}{matching}
    \ForEach{pair of nodes $\textit{(n,m)} \in \textit{N}$} {
    $\{{p_{nm}}^{[r]}\}_{r\in[N]}$ $\leftarrow$ multilateration$\left(\{d_{i,j}\}\right)$;\\
    $transform\left(\{{p_{nm}}^{[r]}\}\right)$;
    }
    \SetKwProg{Fn}{Function}{:}{}
    % \SetKwProg{Fn}{Function}{:}{}
    \Fn{\Fmatch{${\{{p^{[r]}}\}_{\frac{N\times (N-1)}{2}}}$}}{
    $\hspace{1em}{\overline{p}}^{[r]} \leftarrow average\_positions\left(\left\{{p^{[r]}}\right\}_{\frac{N\times (N-1)}{2}}\right)$;\\
    \ForEach{${p_{nm}}^{[r]}$} {
    \SetKwProg{Fn}{Function}{:}{}
    \Fn{\Flse{${{p_{nm}}^{[r]}}$}}{
    $\hspace{1em}{{p}_\Delta}^{[r]} \leftarrow linear\_shift\left({{p_{nm}}}^{[r]}, \Delta X , \Delta Y, \theta\right)$;\\
    $\hspace{1em}{{e}_{nm}} \leftarrow least\_square\_error\left({\overline{p}}^{[r]}, {{p}_\Delta}^{[r]} \right)$;\\
    \If {${{e}_{nm} <  threshold}$} {
        $threshold = {e}_{nm}$;\\
        $final\_{e}_{nm} = {e}_{nm}$;\\
        $p^{[r]} = {{p}_\Delta}^{[r]}$;
        }
    \KwRet $p^{[r]}, \hspace{1em}final\_{e}_{nm}$;
          }
        }
    \KwRet $\left\{p^{[r]}\right\}_{\frac{N\times (N-1)}{2}} \hspace{1em}, \hspace{1em}\left\{final\_{e}_{nm}\right\}_{\frac{N\times (N-1)}{2}}$;
    }
    % $final\_poses = mean\left(\left\{p^{[r]}\right\}_{\frac{N\times (N-1)}{2}}\right)$;
    $pose^{[r]} = mean \left( \left\{ p^{[r]} \right\}_{\frac{N\times (N-1)}{2}} \right)$;

    \end{algorithm}

\begin{algorithm}[t]

    \small
	\caption{Anomaly detection}
	\label{alg:anomaly detection}
	\KwIn{\\
            \hspace{1em}Number of UWB nodes: $\textit{{N}}$;\\
	    % \hspace{1em}UWB ToF Ranges: \textit{{d}_{(i,j)}}\\
             $\hspace{1em}\left\{p^{[r]}\right\}_{\frac{N\times (N-1)}{2}}$;\\
             $\hspace{1em}\left\{final\_{e}_{nm}\right\}_{\frac{N\times (N-1)}{2}}$;\\
	}
        \KwOut{\\
	    \hspace{1em}Anomaly UWB node(s) : anomaly\\
	}
        \ForEach {${i} \in {N}$} {
            \If {${i=n \hspace{1em}\textbf{or}\hspace{1em} i=m}$} {
                ${e_i \hspace{1em}+= final\_{e}_{nm}}$;
        }
            \If {${e_i > threshold}$} {
                $\left\{p^{[r]}\right\}_{\frac{N\times (N-2)}{2}}$ $\leftarrow$ remove$\left({\left\{p^{[r]}\right\}_{\frac{N\times (N-1)}{2}}} , i \right)$;\\
                $sd$ $\leftarrow$ standard\_deviation$\left({\left\{p^{[r]}\right\}_{\frac{N\times (N-1)}{2}}} \right)$;\\
                $\overline{sd}$ $\leftarrow$ standard\_deviation$\left({\left\{p^{[r]}\right\}_{\frac{N\times (N-2)}{2}}} \right)$;\\
                \If{${{sd-\overline{sd}} > threshold}$}{
                
                anomaly=i
                }
        }
    }
\end{algorithm}

\begin{figure*}
    \centering
    \input{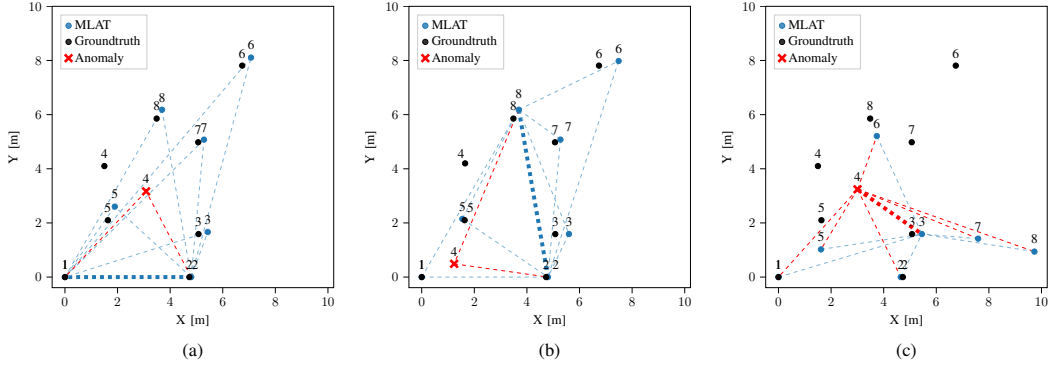}
    \caption{Illustration of node positions calculated using the multilateration method. The blue points represent normal nodes, while the red node generates anomalous ranges. The dotted red lines illustrate noisy UWB range measurements generated by the anomaly node. Subfigures (a) and (b) are configurations generated using two different pairs of normal nodes as the reference basis, while in (c) one of the reference nodes is the anomaly.}
    \label{fig:points_subsets_ml}
\end{figure*}

\subsection{UWB-Based cooperative Localization}

In the remaining portions of this paper, we use the following notation. $N$ robots are considered, with their positions indicated by $p^{i}$ , with $i\in\{{1,..., N\}}$. Based on the time of flight of a message, ToF localization measures the distance between two nodes by exchanging messages between them and can be computed following Eq.~\eqref{eq:tof}:

\begin{equation}
    \label{eq:tof}
    ToF = \frac{{T_{init}}-{T_{res}}}{2}  
\end{equation}

\noindent where \(T_{init}\) is the time measured between the instant when a ranging request is initiated by a given node to another, and until the reply from the latter is received at the former. With \(T_{res}\) we denote the time it takes the receiver to receive, process and return the message. Using this information embedded in the response message, we can determine the distance between the two nodes.

In our proposed cooperative localization method, each point's position $p^{i}$ is determined based on all possible base anchor pairs in the system. Let ${p_{nm}}^{[r]}$ be the layout generated by each base anchor pair ${n}$ and ${m}$. Each target robot's position in this layout is calculated using the triangulation technique in which the first base anchor is assumed to be located at the origin $(0,0)$, and the second one, at a known distance $d$ along the x-axis ${({d}_{(n,m)},0)}$. 

\begin{equation}
    \label{eq:triang_x}
    x_i =   \frac{{d}_{(n,m)}^2 + {d}_{(n,i)}^2 - {d}_{(i,m)}^2}{2\times {d}_{(n,m)}}
\end{equation}

\begin{equation}
    \label{eq:triang_y}
    y_i = \pm{(\sqrt{{d}_{(n,i)}^2 - {x_i}^2})}
\end{equation}

As this technique finds two possible intersection points, according to Eq.~\eqref{eq:triang_y}, the distance between the first target point (robot) and the next points is taken into account to eliminate one of the intersection points based on the distance error. The generated layouts are later transferred based on the two fixed nodes for a common orientation reference. Figure~\ref{fig:points_subsets_ml} shows three different layouts generated by various pair base nodes. 

Once all layouts are generated, the robots' final positions can be defined as shown in Algorithm~\ref{alg:Collabrative localization}. In the first step of the matching function, average positions are used to calculate the minimum least square error (LSE) in shifted layouts. As the objective is to detect the anomaly, we linearly shift the layout in three dimensions $x$, $y$, and $\theta$, to align all configurations. By doing so, the layouts will be robust against errors caused by the base nodes' ranging error. Final positions are estimated based on configurations with the lowest least squares error (LSE).

\begin{figure}[t]
    \input{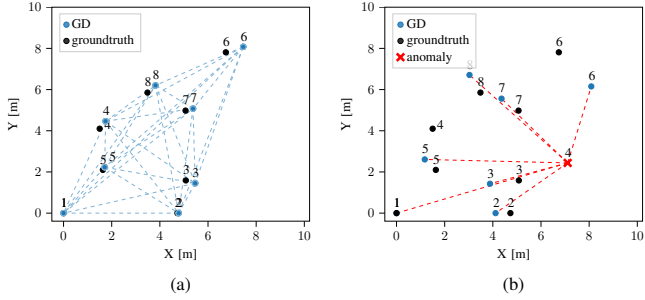}
    \caption{Illustration of calculated positions using gradient descent method in (a) normal nodes, and (b) in the existence of anomaly.}
    \label{fig:points_subsets_gd}
\end{figure}

\subsection{Identification of Anomaly Nodes}

During the cooperative localization step, the final layouts and their LSE were computed. In order to identify anomalous nodes, each individual node's error is determined based on the ${N-1}$ configurations in which it has been involved. Whenever the error of an anchor exceeds a certain threshold, all related configurations to that anchor are removed. Fig.~\ref{fig:points_subsets_ml}\,(c) illustrates how a configuration is developed using an anomaly as a base node. It is possible to identify more false anomalies along with true anomalies when two nodes are involved in developing a wrong configuration. In order to address this, the standard deviation of the positions of each anchor is measured after removing the related configurations of each potential anomaly:

\begin{align}
    \label{eq:std_removed}
   \overline{sd} &= \sum_{i=1}^N{\sqrt{\frac{\sum_{j=1}^{m}{\left({p_j}^i - \mu\right)}}{m}}} & , \quad  m &= \frac{n\times (n-2)}{2}
\end{align}

Where $N$ is the number of UWB nodes. Then the obtained $\overline{sd}$ is compared to the original layouts to assess the effect of each potential anomaly. At this point, it is also apparent that anomalies have the highest dispersion. The anomaly detection process is summarized in Algorithm~\ref{alg:anomaly detection}.

\begin{figure}[t]
    \centering
    \begin{minipage}[t]{.3\textwidth}
      % \centering
      \input{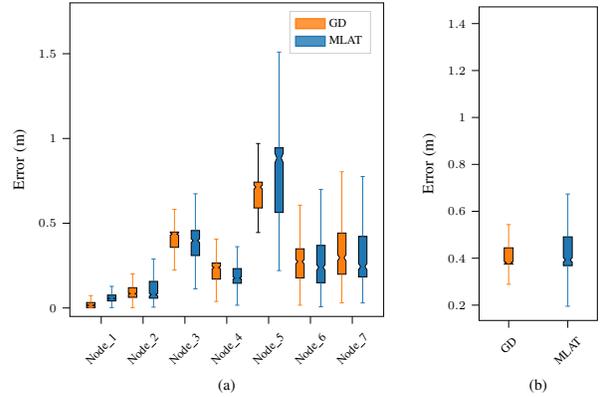}
    \end{minipage}%
    \begin{minipage}[t]{.2\textwidth}
      % \centering
      % This file was created with tikzplotlib v0.10.1.
\begin{tikzpicture}[scale=0.65]

\definecolor{darkgray176}{RGB}{176,176,176}
\definecolor{darkorange25512714}{RGB}{255,127,14}
\definecolor{blue}{RGB}{31,119,180}
\definecolor{red}{RGB}{255,127,14}
\begin{axis}[
tick align=outside,
width=4cm,
height = 8cm,
tick pos=left,
x grid style={darkgray176},
xmin=0.5, xmax=2.5,
xtick style={color=black},
y grid style={darkgray176},
xlabel={\small (b)},
ylabel={Error (m)},
xlabel style = {font=\scriptsize},
ylabel style = {font=\small},
yticklabel style = {font=\scriptsize},
xticklabel style = {font=\scriptsize,rotate=45},
ymin=0.134695075336939, ymax=1.47125730092457,
ytick style={color=black},
xtick={1,2},
xticklabels={GD, MLAT}
]
\path [draw=black, fill=red]
(axis cs:0.925,0.375436586804578)
--(axis cs:1.075,0.375436586804578)
--(axis cs:1.075,0.375555631967057)
--(axis cs:1.0375,0.380658072929989)
--(axis cs:1.075,0.385760513892921)
--(axis cs:1.075,0.444378700285336)
--(axis cs:0.925,0.444378700285336)
--(axis cs:0.925,0.385760513892921)
--(axis cs:0.9625,0.380658072929989)
--(axis cs:0.925,0.375555631967057)
--(axis cs:0.925,0.375436586804578)
--cycle;
\addplot [red]
table {%
1 0.375436586804578
1 0.288859101207334
};
\addplot [red]
table {%
1 0.444378700285336
1 0.543292057154782
};
\addplot [red]
table {%
0.9625 0.288859101207334
1.0375 0.288859101207334
};
\addplot [red]
table {%
0.9625 0.543292057154782
1.0375 0.543292057154782
};
% \addplot [red, mark=+, mark size=2, mark options={solid,fill opacity=0}, only marks]
% table {%
% 1 0.871870678388016
% 1 0.633503184815387
% 1 0.712581269110343
% 1 0.728279153011175
% 1 0.647147814596247
% 1 1.04125806438333
% 1 0.608530507092744
% 1 0.723590677940137
% 1 0.708936897181125
% 1 0.648935047895026
% 1 0.695234694527487
% 1 0.678829629188364
% 1 0.567084157445238
% 1 0.561825411196951
% 1 0.617420960300498
% 1 0.594165252856996
% 1 0.609353628575318
% 1 0.549822171686304
% 1 0.558472049252052
% 1 0.768433100088458
% 1 1.00331505699327
% 1 0.632308919085804
% 1 1.01263492841074
% 1 1.08236000358114
% 1 0.598810538226727
% 1 0.641452054306265
% 1 1.05207117223998
% 1 0.797766661121317
% 1 0.805334713588733
% 1 0.732888910343853
% 1 0.582660122227698
% 1 0.690801822379522
% 1 0.966828179424926
% 1 0.548172025052119
% 1 0.565983967038845
% 1 0.553881213895099
% 1 0.554505575048802
% 1 0.769632556508827
% 1 1.11481883961634
% 1 1.05866762033662
% 1 0.646400047766696
% 1 0.668110188071791
% 1 0.741786038322579
% 1 0.652145008434283
% 1 0.688999443490605
% 1 0.840893610201757
% 1 0.795937110743129
% 1 0.642591550497605
% 1 0.568365182088394
% };
\path [draw=black, fill=blue]
(axis cs:1.925,0.368343714059752)
--(axis cs:2.075,0.368343714059752)
--(axis cs:2.075,0.383758949585357)
--(axis cs:2.0375,0.392841969477669)
--(axis cs:2.075,0.401924989369982)
--(axis cs:2.075,0.491069796088535)
--(axis cs:1.925,0.491069796088535)
--(axis cs:1.925,0.401924989369982)
--(axis cs:1.9625,0.392841969477669)
--(axis cs:1.925,0.383758949585357)
--(axis cs:1.925,0.368343714059752)
--cycle;
\addplot [blue]
table {%
2 0.368343714059752
2 0.195447903772741
};
\addplot [blue]
table {%
2 0.491069796088535
2 0.673496607677857
};
\addplot [blue]
table {%
1.9625 0.195447903772741
2.0375 0.195447903772741
};
\addplot [blue]
table {%
1.9625 0.673496607677857
2.0375 0.673496607677857
};
% \addplot [blue, mark=+, mark size=2, mark options={solid,fill opacity=0}, only marks]
% table {%
% 2 0.777954945498336
% 2 0.682182101395221
% 2 0.770831920229417
% 2 0.906098617142774
% 2 0.770027787674679
% 2 0.932192566337355
% 2 0.855975791833064
% 2 0.871126274032847
% 2 0.940238293985858
% 2 0.995699196241419
% 2 0.956751640248456
% 2 0.789632448626225
% 2 0.926571542354851
% 2 0.98265328048677
% 2 0.706417383734118
% 2 0.772066115050842
% 2 1.2312845729414
% 2 1.01553855127841
% 2 0.994381582204159
% 2 0.894696026207911
% 2 1.30513813129525
% 2 0.84412313327373
% 2 1.22014466094246
% 2 1.41050447248877
% 2 0.884364947012756
% 2 0.725744539577079
% 2 0.763288711020872
% 2 0.741373725185875
% 2 0.894200644531006
% 2 1.405174493142
% 2 1.37272198040162
% 2 1.13361159040726
% };
\addplot [darkorange25512714]
table {%
0.9625 0.380658072929989
1.0375 0.380658072929989
};
\addplot [blue]
table {%
1.9625 0.392841969477669
2.0375 0.392841969477669
};
\end{axis}

\end{tikzpicture}
    \end{minipage}
    \caption{Multilateration and gradient descent-based trajectory error for (a) each individual node, and (b) the average error for normal dataset. The error distribution is significantly larger in Node\,5 due to repeated NLOS conditions during the experiments.}
    \label{fig:true_error}
\end{figure}

\begin{figure*}[t]
    \centering
    \input{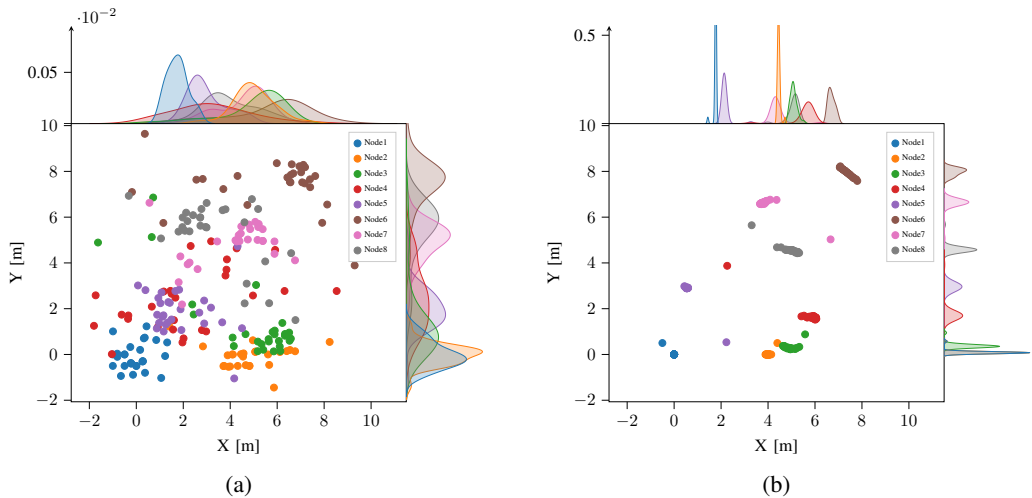}
    \caption{Example of estimated positions, from a single set of measurements,  using MLAN (left) and GD (right). The data includes anomalies. Each of the data points represents the position calculated using a single pair of nodes for the origin and x-axis orientation. All the estimates are then rotated and shifted as needed for comparison.}
    \label{fig:jointpoints subsets mln}
\end{figure*}

\subsection{Gradient Descent Optimization}

Gradient descent as an optimization algorithm has been used to reduce the loss function in various models. The goal of gradient descent in UWB localization is to position the nodes by minimizing the distance error~\cite{ridolfi2021uwb}. The basic idea behind gradient descent is to iteratively update the positions by taking a step in the opposite direction of the gradient. By iterating this process, gradient descent gradually converges to final positions that minimize the error function. Gradient descent requires initial parameter values to start the optimization process. For the convergence speed and stability of the algorithm, the initial reference system for the optimization process has been calculated using the UWB collaborative localization method described in subsection\,B. After the anchors are initialized, their positions are modified by minimizing the loss function. The loss function is defined as the difference between points $p^{i}$ and $p^{j}$ reported UWB distance ${d}_{(i,j)}$, and their estimated 2D distance~\eqref{eq:gd_error}:

\begin{equation}
    \label{eq:gd_error}
   Loss = \sum_{i,j=1}^N {\left({{d}_{(i,j)}} - {\lVert {{p^{i}} - {p^{j}}}\rVert }\right)}^2
\end{equation}

Figure~\ref{fig:points_subsets_gd} illustrates how each point coordinates with other nodes to be located at an optimal position based on the loss function. In Fig.~\ref{fig:points_subsets_gd}\,(b), it is shown that with the gradient descent method, anomalies impact also other nodes' positions regardless of the base nodes, as the global error is minimized, with anomalies generating a non-convex cost function.

\section{Experimental Results}\label{sec:experimental_results}

\begin{figure*}[t]
    % \centering
    \input{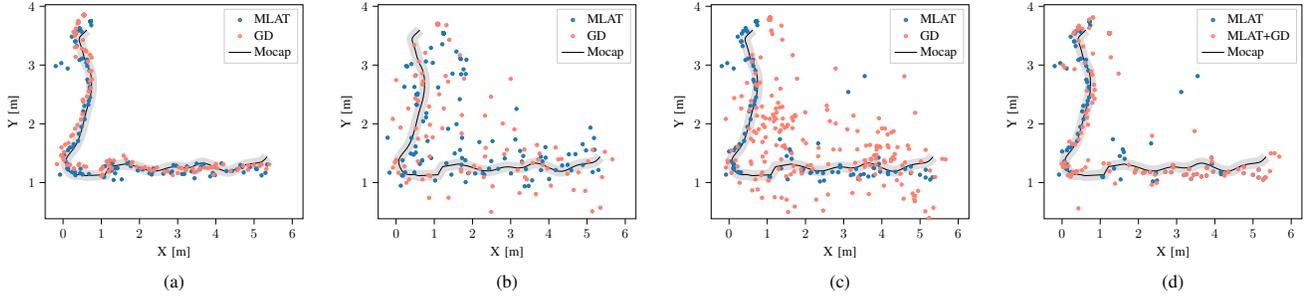}
    \caption{Estimated trajectories of a normal agent (node) using multilateration and gradient descent methods in the (a) absence, (b) presence of an anomalous node, (c) after removing the anomalous layout by each method, and (d) after removing the anomaly and optimizing the positions by gradient descent.}
    \label{fig:trajectory_true}
\end{figure*}

In this section, we focus on the results of experiments carried out to validate the functionality of the proposed methods. First, we evaluate the viability of the proposed cooperative localization framework on normal UWB ranges in comparison to the optimized positions using the gradient descent algorithm. Then, we assess the behavior of these methods in the presence of anomalies.

\subsection{cooperative Localization with Nominal Ranging}

The findings of this study regarding the trajectory error of the multilateration method based on the anchor-free cooperative framework for normal data for different agents are presented in Fig.~\ref{fig:true_error}. Specifically, the performance of the multilateration method was evaluated and compared with the gradient descent optimization technique. Through the implementation of gradient descent, the position of each anchor was optimized, which resulted in a slight reduction in the total error as compared to the multilateration method. However, it is worth noting that the computational time required by the gradient descent optimization technique is higher than that of the multilateration method. Therefore, while the gradient descent optimization technique may offer a slight improvement in positioning accuracy, it may not be the most practical approach in scenarios where computational efficiency is a critical consideration. As an example, Figure~\ref{fig:trajectory_true}\,(a) provides the trajectory results of both methods for the normal dataset. During the data acquisition, we wanted to study whether the anomaly would be detected independently of having LOS or NLOS ranging conditions. Therefore, the trajectories of the nodes are generated in a way such that Node\,5 range measurements suffer of NLOS conditions significantly more than others.

% \begin{figure}[t]
% \centering
% \begin{minipage}[t]{.3\textwidth}
%   % \centering
%   \input{tex/bx_traj.tex}
% \end{minipage}%
% \begin{minipage}[t]{.2\textwidth}
%   % \centering
%   \input{tex/bx_points.tex}
% \end{minipage}
% \caption{Multilateration and gradient descent positioning error(true).}
% \label{fig:true_error}
% \end{figure}

% \begin{figure*}
%     \centering
%     \input{tex/jointpoints_12_ml_anom.tex}
%     \caption{Example of estimated positions, from a single set of measurements,  using MLAN (left) and GD (right). The data includes anomalies. Each of the data points represents the position calculated using a single pair of nodes for the origin and x-axis orientation. \red{All the estimates are then rotated and mirrored as needed for comparison.}}
%     \label{fig:jointpoints subsets mln}
% \end{figure*}

\subsection{Anomaly Detection}

The multilateration method proposed in this study utilizes linear transformations and shifts for all configurations. As a result, the estimated distances between nodes remain fixed at each timestamp, enabling the use of dispersion-based anomaly detection techniques to detect any irregular layouts or nodes. Figure~\ref{fig:jointpoints subsets mln} presents a clear visualization of the node distribution at a specific timestamp. It is evident from subfigure (a) that the anomalous node was transmitting erroneous UWB ranges to all nodes, resulting in a wide distribution of positions.

In the context of detecting anomalies, gradient descent minimizes the distance error between the predicted values and the actual values of each node. As a result, the positions of each node from different configurations have converged in a way that leads to similar distributions for all nodes, including the anomaly node. This means that, at each timestamp, the anomaly node may not stand out as being significantly different from the other nodes, as shown in subfigure (b). Figure~\ref{fig:trajectory_true}.(a) and (b) illustrate the calculated trajectory of a normal node in the absence of an anomaly and disturbed by an anomaly, respectively, using both the proposed multilateration and gradient descent techniques. In Subfigure (c), anomaly detection and elimination results are shown using both methods, while subfigure (d) shows slightly different results when multilateration anomaly detection is followed by gradient descent optimization for positioning.

% \begin{figure*}
%     \centering
%     \input{tex/6traj_anom.tex}
%     \caption{Trajectories of the different robots during the experiments(with anomaly).}
%     \label{fig:trajectory_anom}
% \end{figure*}

\begin{figure}[t]
\centering
\begin{minipage}[t]{.3\textwidth}
  % \centering
  \input{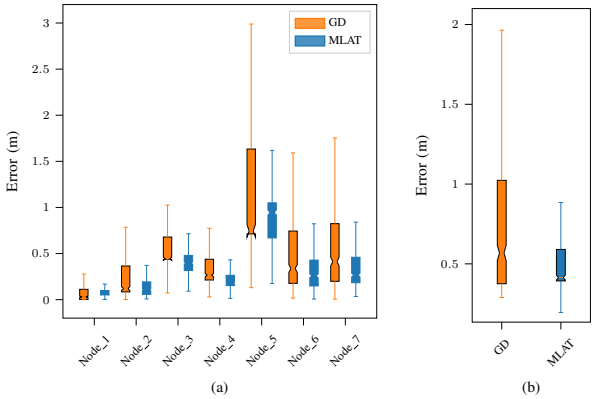}
\end{minipage}%
\begin{minipage}[t]{.2\textwidth}
  % \centering
  % This file was created with tikzplotlib v0.10.1.
\begin{tikzpicture}[scale=0.65]

\definecolor{darkgray176}{RGB}{176,176,176}
\definecolor{darkorange25512714}{RGB}{255,127,14}
\definecolor{blue}{RGB}{31,119,180}
\definecolor{red}{RGB}{255,127,14}
\begin{axis}[
tick align=outside,
width=4cm,
height = 8cm,
tick pos=left,
x grid style={darkgray176},
xmin=0.5, xmax=2.5,
xtick style={color=black},
y grid style={darkgray176},
xlabel={\small (b)},
ylabel={Error (m)},
xlabel style = {font=\scriptsize},
ylabel style = {font=\small},
yticklabel style = {font=\scriptsize},
xticklabel style = {font=\scriptsize,rotate=45},
ymin=0.134695075336939, ymax=2.1,
ytick style={color=black},
xtick={1,2},
xticklabels={GD, MLAT}
]
\path [draw=black, fill=red]
(axis cs:0.925,0.375436586804578)
--(axis cs:1.075,0.375436586804578)
--(axis cs:1.075,0.520757749194583)
--(axis cs:1.0375,0.56871669267675)
--(axis cs:1.075,0.616675636158916)
--(axis cs:1.075,1.02343838594054)
--(axis cs:0.925,1.02343838594054)
--(axis cs:0.925,0.616675636158916)
--(axis cs:0.9625,0.56871669267675)
--(axis cs:0.925,0.520757749194583)
--(axis cs:0.925,0.375436586804578)
--cycle;
\addplot [red]
table {%
1 0.375436586804578
1 0.288859101207334
};
\addplot [red]
table {%
1 1.02343838594054
1 1.96437358049734
};
\addplot [red]
table {%
0.9625 0.288859101207334
1.0375 0.288859101207334
};
\addplot [red]
table {%
0.9625 1.96437358049734
1.0375 1.96437358049734
};
% \addplot [red, mark=+, mark size=2, mark options={solid,fill opacity=0}, only marks]
% table {%
% 1 2.37521011564713
% 1 2.16984521421877
% 1 2.16654293657078
% 1 1.99642739197263
% 1 2.3898374337046
% };
\path [draw=black, fill=blue]
(axis cs:1.925,0.392841969477669)
--(axis cs:2.075,0.392841969477669)
--(axis cs:2.075,0.399460845538047)
--(axis cs:2.0375,0.414085239390405)
--(axis cs:2.075,0.428709633242763)
--(axis cs:2.075,0.590440849676622)
--(axis cs:1.925,0.590440849676622)
--(axis cs:1.925,0.428709633242763)
--(axis cs:1.9625,0.414085239390405)
--(axis cs:1.925,0.399460845538047)
--(axis cs:1.925,0.392841969477669)
--cycle;
\addplot [blue]
table {%
2 0.392841969477669
2 0.195447903772741
};
\addplot [blue]
table {%
2 0.590440849676622
2 0.884364947012756
};
\addplot [blue]
table {%
1.9625 0.195447903772741
2.0375 0.195447903772741
};
\addplot [blue]
table {%
1.9625 0.884364947012756
2.0375 0.884364947012756
};
% \addplot [blue, mark=+, mark size=2, mark options={solid,fill opacity=0}, only marks]
% table {%
% 2 0.895693362383832
% 2 1.09993952891632
% 2 0.896197276595678
% 2 1.08122501090777
% 2 1.3946847907626
% 2 0.977299215209203
% 2 0.9246941606225
% 2 0.902532403939736
% 2 1.0959592302719
% 2 1.02986285291149
% 2 0.890456126808796
% 2 0.891668099599189
% 2 1.7302131180638
% 2 1.22014466094246
% 2 1.41050447248877
% 2 0.894200644531006
% 2 1.405174493142
% 2 1.37272198040162
% 2 1.13361159040726
% };
\addplot [darkorange25512714]
table {%
0.9625 0.56871669267675
1.0375 0.56871669267675
};
\addplot [darkorange25512714]
table {%
1.9625 0.414085239390405
2.0375 0.414085239390405
};
\end{axis}

\end{tikzpicture}
\end{minipage}
\caption{Multilateration and gradient descent-based trajectory error for (a) each individual node, and (b) the average error in the presence of an anomaly. The errors increase in a mostly linear relation with respect to the anomaly-free data, for both LOS and NLOS (Node 5) measurements.}
\label{fig:anom_error}
\end{figure}

\begin{figure}[t]
% \centering
\begin{minipage}[t]{.23\textwidth}
  % \centering
  % This file was created with tikzplotlib v0.10.1.
\begin{tikzpicture}[scale=0.85]

\definecolor{darkgray176}{RGB}{176,176,176}

\begin{axis}[
% colorbar,
% colorbar style={ylabel={}},
% colormap={mymap}{[1pt]
%   rgb(0pt)=(0.968627450980392,0.984313725490196,1);
%   rgb(1pt)=(0.870588235294118,0.92156862745098,0.968627450980392);
%   rgb(2pt)=(0.776470588235294,0.858823529411765,0.937254901960784);
%   rgb(3pt)=(0.619607843137255,0.792156862745098,0.882352941176471);
%   rgb(4pt)=(0.419607843137255,0.682352941176471,0.83921568627451);
%   rgb(5pt)=(0.258823529411765,0.572549019607843,0.776470588235294);
%   rgb(6pt)=(0.129411764705882,0.443137254901961,0.709803921568627);
%   rgb(7pt)=(0.0313725490196078,0.317647058823529,0.611764705882353);
%   rgb(8pt)=(0.0313725490196078,0.188235294117647,0.419607843137255)
% },
point meta max=12,
point meta min=0,
tick align=outside,
tick pos=left,
width=5cm,
height = 5cm,
xlabel style = {font=\scriptsize , align=center},
ylabel style = {font=\scriptsize, align=center},
yticklabel style = {font=\tiny, rotate=45.0},
x grid style={darkgray176},
xlabel=Predicted label\\\\ {\small (a) Gradient Descent},
xmin=-0.5, xmax=2.5,
xtick style={color=black},
xtick={0,1,2},
xticklabel style={font=\tiny, rotate=45.0},
xticklabels={Byzantine,Other,None},
y dir=reverse,
y grid style={darkgray176},
ylabel={True label},
ymin=-0.5, ymax=2.5,
ytick style={color=black},
ytick={0,1,2},
yticklabels={Byzantine,Other,None}
]
\addplot graphics [includegraphics cmd=\pgfimage,xmin=-0.5, xmax=2.5, ymin=2.5, ymax=-0.5] {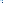};
\draw (axis cs:0,0) node[
  scale=0.75,
  anchor=base,
  text=black,
  rotate=0.0
]{12 \%};
\draw (axis cs:1,0) node[
  scale=0.75,
  anchor=base,
  text=black,
  rotate=0.0
]{8.0 \%};
\draw (axis cs:2,0) node[
  scale=0.75,
  anchor=base,
  text=black,
  rotate=0.0
]{29 \%};
\draw (axis cs:0,1) node[
  scale=0.75,
  anchor=base,
  text=black,
  rotate=0.0
]{0.0};
\draw (axis cs:1,1) node[
  scale=0.75,
  anchor=base,
  text=black,
  rotate=0.0
]{0.0};
\draw (axis cs:2,1) node[
  scale=0.75,
  anchor=base,
  text=black,
  rotate=0.0
]{0.0};
\draw (axis cs:0,2) node[
  scale=0.75,
  anchor=base,
  text=black,
  rotate=0.0
]{0.0};
\draw (axis cs:1,2) node[
  scale=0.75,
  anchor=base,
  text=black,
  rotate=0.0
]{0.0};
\draw (axis cs:2,2) node[
  scale=0.75,
  anchor=base,
  text=white,
  rotate=0.0
]{50 \%};
\end{axis}

\end{tikzpicture}
\end{minipage}%
\begin{minipage}[t]{.2\textwidth}
  % \centering
  % This file was created with tikzplotlib v0.10.1.
\begin{tikzpicture}[scale=0.85]

\definecolor{darkgray176}{RGB}{176,176,176}

\begin{axis}[
% colorbar,
% colorbar style={ylabel={}},
% colormap={mymap}{[1pt]
%   rgb(0pt)=(0.968627450980392,0.984313725490196,1);
%   rgb(1pt)=(0.870588235294118,0.92156862745098,0.968627450980392);
%   rgb(2pt)=(0.776470588235294,0.858823529411765,0.937254901960784);
%   rgb(3pt)=(0.619607843137255,0.792156862745098,0.882352941176471);
%   rgb(4pt)=(0.419607843137255,0.682352941176471,0.83921568627451);
%   rgb(5pt)=(0.258823529411765,0.572549019607843,0.776470588235294);
%   rgb(6pt)=(0.129411764705882,0.443137254901961,0.709803921568627);
%   rgb(7pt)=(0.0313725490196078,0.317647058823529,0.611764705882353);
%   rgb(8pt)=(0.0313725490196078,0.188235294117647,0.419607843137255)
% },
point meta max=12,
point meta min=0,
tick align=outside,
tick pos=left,
width=5cm,
height = 5cm,
xlabel style = {font=\scriptsize , align=center},
ylabel style = {font=\scriptsize, align=center},
yticklabel style = {font=\tiny, rotate=45.0},
x grid style={darkgray176},
xlabel= Predicted label\\\\ {\small (b) Multilateration},
xmin=-0.5, xmax=2.5,
xtick style={color=black},
xtick={0,1,2},
xticklabel style={font=\tiny, rotate=45.0},
xticklabels={Byzantine,Other,None},
y dir=reverse,
y grid style={darkgray176},
ylabel={True label},
ymin=-0.5, ymax=2.5,
ytick style={color=black},
ytick={0,1,2},
yticklabels={Byzantine,Other,None}
]
\addplot graphics [includegraphics cmd=\pgfimage,xmin=-0.5, xmax=2.5, ymin=2.5, ymax=-0.5] {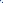};
\draw (axis cs:0,0) node[
  scale=0.75,
  anchor=base,
  text=white,
  rotate=0.0
]{46 \%};
\draw (axis cs:1,0) node[
  scale=0.75,
  anchor=base,
  text=black,
  rotate=0.0
]{0.0};
\draw (axis cs:2,0) node[
  scale=0.75,
  anchor=base,
  text=black,
  rotate=0.0
]{4.0 \%};
\draw (axis cs:0,1) node[
  scale=0.75,
  anchor=base,
  text=black,
  rotate=0.0
]{0.0};
\draw (axis cs:1,1) node[
  scale=0.75,
  anchor=base,
  text=black,
  rotate=0.0
]{0.0};
\draw (axis cs:2,1) node[
  scale=0.75,
  anchor=base,
  text=black,
  rotate=0.0
]{0.0};
\draw (axis cs:0,2) node[
  scale=0.75,
  anchor=base,
  text=black,
  rotate=0.0
]{0.0};
\draw (axis cs:1,2) node[
  scale=0.75,
  anchor=base,
  text=black,
  rotate=0.0
]{0.0};
\draw (axis cs:2,2) node[
  scale=0.75,
  anchor=base,
  text=white,
  rotate=0.0
]{50 \%};
\end{axis}

\end{tikzpicture}
\end{minipage}
\caption{Confusion matrix for the two anomaly detection methods studied in this paper. The optimization done in the gradient descent method hinders the detection of anomalies owing to a lower global relative positioning error.}
\label{fig:Confusion_matrix}
\end{figure}
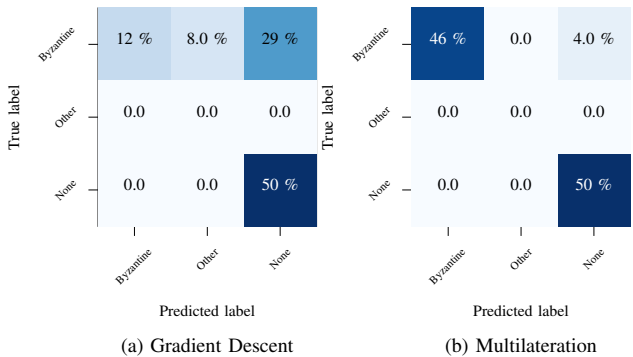

Figure~\ref{fig:anom_error} also displays the trajectory error of all agents for both methods. Notably, in the case of anomalies, the gradient descent algorithm yields a higher error at each node and also results in a larger total error.

In the confusion matrix, in figure~\ref{fig:Confusion_matrix}, we evaluated the performance of both methods with and without the anomaly. As previously discussed, both techniques perform similarly, with a balance between speed and accuracy when dealing with normal data, marked as None in the matrix. However, when it comes to detecting anomalies, multilateration provides better accuracy, and the false negative rate of the gradient descent technique is high, indicating that the model is missing a significant number of anomalies.
\section{Conclusion and Future Work}\label{sec:conclusion}

This paper proposes a novel approach to detecting anomalies in UWB ranging in mobile robots by exploiting redundant localization data. Specifically, throughout the paper, we have studied the problem of identifying anomalous nodes in a multi-robot system. The anomalies can be caused by byzantine agents, due to either malicious behavior or hardware or timing problems. Both multilateration and optimization-based methods are typical in the literature in terms of relative localization estimation. Despite the higher accuracy of optimization methods, we show in this study how leveraging redundancy allows for multilateration methods to be more robust in identifying anomalies. Our approach consists in calculating relative positions with different subsets of ranging data, with results that can then be compared to each other. In comparison, a gradient descent method analyzed in the paper that minimizes global error masks the anomalies instead. Once the anomalous nodes are detected, however, optimization-based methods can be used to increase the accuracy of the relative localization. Overall, our experiments with multiple real robots show that the proposed method works and clearly outperforms anomaly detection based on a gradient descent method for relative localization.

In future work, we will study different types of anomalies. We are particularly interested in analyzing the performance of this method for NLOS ranging detection, specially when only a subset of ranges from a given node present anomalies.

%%%%%%%%%%%%%%%%%%%%%%%%%%%%%%%%%%%%%%%%%%%%%%
%%                                          %%
%%            ACKNOWLEDGMENT                %%
%%                                          %%
%%%%%%%%%%%%%%%%%%%%%%%%%%%%%%%%%%%%%%%%%%%%%%

% \newpage

\section*{Acknowledgment}

This research work is supported by the Academy of Finland's AeroPolis project (Grant No. 348480), and by the R3Swarms project funded by the Secure Systems Research Center (SSRC), Technology Innovation Institute (TII).

%%%%%%%%%%%%%%%%%%%%%%%%%%%%%%%%%%%%%%%%%%%%%%
%%                                          %%
%%              BIBLIOGRAPHY                %%
%%                                          %%
%%%%%%%%%%%%%%%%%%%%%%%%%%%%%%%%%%%%%%%%%%%%%%

% \newpage

\bibliographystyle{unsrt}
\bibliography{bibliography}
\end{document}